%% file: semtagger.tex
\documentclass[11pt]{article}
\usepackage{coling2016}
\usepackage{times}
\usepackage{url}
\usepackage{latexsym}
\usepackage{amssymb}
\usepackage{amsmath}
\usepackage{booktabs}

\usepackage{covington}
\usepackage{graphicx}

\title{Semantic Tagging with Deep Residual Networks}

\author{Johannes Bjerva \\
  University of Groningen \\
  The Netherlands \\
  {\tt j.bjerva@rug.nl} \\\And
  Barbara Plank \\
  University of Groningen \\
  The Netherlands \\
  {\tt b.plank@rug.nl} \\\And
  Johan Bos \\
  University of Groningen \\
  The Netherlands \\
  {\tt johan.bos@rug.nl} \\}

\date{}

\begin{document}
\maketitle
\begin{abstract}  
We propose a novel semantic tagging task, \textit{sem-tagging}, tailored for the purpose of multilingual semantic parsing,
and present the first tagger using deep residual networks (ResNets).
Our tagger uses both word and character representations, and includes a novel residual bypass architecture.
We evaluate the tagset both intrinsically on the new task of semantic tagging, as well as on Part-of-Speech (POS) tagging. 
Our system, consisting of a ResNet and an auxiliary loss function predicting our semantic tags, significantly outperforms prior results on English Universal Dependencies POS tagging (95.71\% accuracy on UD v1.2 and 95.67\% accuracy on UD v1.3).
\end{abstract}

\blfootnote{
    %
    %
    %
    %
    \hspace{-0.65cm}  
    This work is licensed under a Creative Commons
    Attribution 4.0 International Licence.
    Licence details:
    \url{http://creativecommons.org/licenses/by/4.0/}
    %
    %
}

\input{intro}

\input{semantic_tagging}

\input{method}

\input{model_description}
\input{experiments}

\input{discussion}

\input{conclusions}

\section*{Acknowledgements}
The authors would like to thank Robert \"Ostling for tips on ResNets,
and Calle B\"orstell, Johan Sjons, Lasha Abzianidze, and the anonymous reviewers for feedback on earlier versions of this manuscript.
We would like to thank the Center for Information Technology of the University of Groningen for their support
and for providing access to the Peregrine high performance computing cluster.
This work was partially funded by the NWO--VICI grant \mbox{``Lost in Translation -- Found in Meaning''} (288-89-003).

\bibliographystyle{acl}
\bibliography{comsem}

\end{document}

%% file: intro.tex
\section{Introduction}


A key issue in computational semantics is the transferability of
semantic information across languages. Many semantic parsing systems
depend on sources of information such as POS tags
\cite{pradhan:2004,mrs,step:boxer,Butler:2010,berant:2014}. However,
these tags are often customised for the language at hand \cite{ptb} or
massively abstracted, such as the Universal Dependencies tagset
\cite{nivre:2016}. Furthermore, POS tags are syntactically oriented,
and therefore often contain both irrelevant and insufficient
information for semantic analysis and deeper semantic processing. This
means that, although POS tags are highly useful for many downstream
tasks, they are unsuitable both for semantic parsing in general, and
for tasks such as recognising textual entailment.

We present a novel set of semantic labels tailored for the purpose of
multilingual semantic parsing.  This tagset (i) abstracts over
POS and named entity types; (ii) fills gaps in semantic
modelling by adding new categories (for instance for phenomena like
negation, modality, and quantification); and (iii) generalises over
specific languages (see Section~\ref{sec:semtagging}).  We introduce and motivate this new task in this
paper, and refer to it as \textit{semantic tagging}. Our experiments aim to answer the following
two research questions:

\begin{enumerate}

    \item Given an annotated corpus of semantic tags, it is
    straightforward to apply off-the-shelf sequence taggers. Can we
    significantly outperform these with recent neural network
    architectures?

     \item Semantic tagging is essential for deep semantic parsing. Can we find evidence that semtags are effective also for other NLP tasks?

\end{enumerate}

To address the first question, we will look at convolutional neural
networks (CNNs) and recurrent neural networks (RNNs), which are both
highly prominent approaches in the recent natural language processing
(NLP) literature.  A recent development is the emergence of deep
residual networks (ResNets), a building block for CNNs.  ResNets
consist of several stacked residual units, which can be thought of as
a collection of convolutional layers coupled with a `shortcut' 
which aids the propagation of the signal in a neural network.  This
allows for the construction of much deeper networks, since keeping a
`clean' information path in the network facilitates optimisation
\cite{resnets:2016}.  ResNets have recently shown state-of-the-art
performance for image classification tasks
\cite{resnets:2015,resnets:2016}, and have also seen some recent use
in NLP \cite{robert:sigmorphon:2016,conneau:2016,bjerva:2016:dsl,google:nmt}.  However, no
previous work has attempted to apply ResNets to NLP tagging tasks.
%

To answer our second question, we carry out an extrinsic evaluation
exercise. We investigate the effect of using semantic tags as an
auxiliary loss for POS tagging. Since POS tags are useful for many NLP
tasks, it follows that semantic tags must be useful if they can
improve POS tagging.

%% file: semantic_tagging.tex
\section{Semantic Tagging}
\label{sec:semtagging}

\subsection{Background}

We refer to \textit{semantic tagging}, or \textit{sem-tagging}, as the
task of assigning semantic class categories to the smallest meaningful
units in a sentence. In the context of this paper these units can be
morphemes, words, punctuation, or multi-word expressions.  The linguistic
information traditionally obtained for deep processing is insufficient
for fine-grained lexical semantic analysis. The widely used Penn
Treebank (PTB) Part-of-Speech tagset \cite{ptb} does not make the
necessary semantic distinctions, in addition to containing redundant
information for semantic processing. Let us consider a couple of
examples.

There are significant differences in meaning between the determiners
\textit{every} (universal quantification), \textit{no} (negation), and
\textit{some} (existential quantification), but they all receive the
\texttt{DT} (determiner) POS label in PTB. Since determiners form a closed
class, 
one could enumerate all word forms for each
class. Indeed some recent implementations of semantic parsing
follow this strategy \cite{step:boxer,Butler:2010}. This might work
for a single language, but it falls short when considering a
multilingual setting. Furthermore, determiners like \textit{any} can
have several interpretations and need to be disambiguated in context.

Semantic tagging does not only apply to determiners, but reaches all
parts of speech. Other examples where semantic classes disambiguate
are reflexive versus emphasising pronouns (both POS-tagged as
\texttt{PRP}, personal pronoun); the comma, that could be a
conjunction, disjunction, or apposition; intersective vs.\ subsective
and privative adjectives (all POS-tagged as \texttt{JJ}, adjective);
proximal vs.\ medial and distal demonstratives (see Example 1); subordinate vs.
coordinate discourse relations; role nouns vs. entity nouns. The set
of semantic tags that we use in this paper is established in a
data-driven manner, considering four languages in a parallel corpus
(English, German, Dutch and Italian). This first inventory of classes
comprises 13 coarse-grained tags and 75 fine-grained tags (see
Table~\ref{table:classes}). As can be seen from this table and the examples given below, the tagset
also includes named entity classes (see also Example 2).

\begin{examples}
    \item{\gll These cats live in that house .
\texttt{PRX} \texttt{CON} \texttt{ENS} \texttt{REL} \texttt{DST} \texttt{CON} \texttt{NIL}
\glt\vspace{-0.35cm}\glend}
    \item{\gll Ukraine 's glory has not yet perished , neither her freedom .
\texttt{GPE} \texttt{HAS} \texttt{CON} \texttt{ENT} \texttt{NOT} \texttt{IST} \texttt{EXT} \texttt{NIL} \texttt{NOT} \texttt{HAS} \texttt{CON} \texttt{NIL}
\glt\vspace{-0.35cm}\glend}
\end{examples}

In Example 1, both \textit{these} and \textit{that} would be tagged as \texttt{DT}.
However, with our semantic tagset, they are disambiguated as \texttt{PRX} (proximal) and \texttt{DST} (distal).
In Example 2, \textit{Ukraine} is tagged as \texttt{GPE} rather than \texttt{NNP}.

\begin{table}[htbp]
\setlength{\tabcolsep}{3pt}
\renewcommand{\arraystretch}{0.7}
\small \begin{tt} 
\begin{tabular}{|p{5mm}p{5mm}p{35mm}|}
  \hline
  ANA & PRO & pronoun\\
      & DEF & definite\\
  &HAS & possessive\\
  &REF & reflexive\\
  &EMP & emphasizing\\
  \hline
  ACT & GRE & greeting\\
    &ITJ & interjection\\
  &HES & hesitation \\
  &QUE & interrogative\\
  \hline
  ATT &QUA &quantity\\
  &UOM &measurement\\
  &IST &intersective\\
  &REL &relation \\
  &RLI &rel.~inv.~scope \\
  &SST &subsective\\
  &PRI &privative \\
  &INT &intensifier\\
  &SCO &score\\
  \hline
  LOG& ALT &alternative\\
 & EXC &exclusive\\
 & NIL &empty \\
 & DIS &disjunct./exist.\\
 & IMP &implication \\
 & AND &conjunct./univ.\\
 & BUT &contrast \\
\hline
\end{tabular}
\begin{tabular}{|p{5mm}p{5mm}p{35mm}|}
  \hline
  COM& EQA &equative\\
 & MOR & comparative~pos.\\
 & LES & comparative~neg.\\
 & TOP & pos.~superlative \\
 & BOT & neg.~superlative\\
 & ORD & ordinal\\
\hline
  DEM& PRX &proximal\\
 & MED& medial \\
 & DST& distal \\
 \hline
  DIS &SUB &subordinate\\
 & COO &coordinate  \\
 & APP &appositional  \\
 \hline
 MOD &NOT &negation \\
 & NEC &necessity \\
 & POS &possibility\\
 \hline
  ENT& CON &concept\\
 & ROL &role \\
 \hline
  NAM &GPE &geo-political~ent.\\
 & PER & person \\
 & LOC &location \\
 & ORG &organisation \\
 & ART & artifact \\
 & NAT &natural~obj./phen.\\
 & HAP & happening \\
 & URL & url\\
  \hline
\end{tabular}
\begin{tabular}{|p{5mm}p{5mm}p{35mm}|}
 \hline
  EVE& EXS & untensed~simple\\
 & ENS &present~simple\\
 & EPS &past~simple\\
 & EFS &future~simple \\
 & EXG &untensed~prog.\\
 & ENG &present~prog.\\
 & EPG &past~prog.\\
 & EFG &future~prog.\\
 & EXT &untensed~perfect\\
 & ENT &present~perfect\\
 & EPT &past~perfect\\
 & EFT &future~perfect \\
 & ETG &perfect~prog.\\
 & ETV &perfect~passive \\
 & EXV &passive \\
 \hline
  TNS& NOW &present~tense \\
 & PST &past~tense \\
 & FUT &future~tense \\
  \hline
  TIM &DOM &day~of~month\\
 & YOC &year~of~century \\
 & DOW &day~of~week\\
 & MOY &month~of~year \\
 & DEC &decade\\
 & CLO &clocktime\\
 & & \\
 & & \\
   \hline
\end{tabular}
\end{tt}
\caption{Semantic tags used in this paper.\label{table:classes}}
\end{table}

\subsection{Annotated data}

We use two semtag datasets. The Groningen Meaning Bank (GMB) corpus of
English texts (1.4 million words) containing silver standard semantic
tags obtained by running a simple rule-based semantic tagger
\cite{gmb:hla}. This tagger uses POS and named entity tags available
in the GMB (automatically obtained with the C\&C tools
\cite{candcboxer:2007} and then manually corrected), as well as a set
of manually crafted rules to output semantic tags. Some tags related
to specific phenomena were hand-corrected in a second stage.

Our second dataset is smaller but equipped with gold standard semantic
tags and used for testing (PMB, the Parallel Meaning Bank). It
comprises a selection of 400 sentences of the English part of a
parallel corpus. It has no overlap with the GMB corpus.  For this
dataset, we used the Elephant tokeniser, which performs word,
multi-word and sentence segmentation \cite{elephant}. We then used the
simple rule-based semantic tagger described above to get an initial
set of tags. These tags were then corrected by a human annotator (one
of the authors of this paper).

For the extrinsic evaluation, we use the POS annotation in the English portion of the Universal Dependencies dataset, version 1.2 and 1.3~\cite{nivre:2016}.
An overview of the data used is shown in Table~\ref{tab:data}.

\begin{table}[h!]
    \small
\begin{center}
\begin{tabular}{lrrrr}
\toprule
\sc Corpus                               & \sc Train (sents/toks)  & \sc Dev (sents/toks) & \sc Test (sents/toks) & \sc n tags \\
\midrule
ST Silver (GMB) & 42,599 / 930,201 & 6,084 / 131,337 & 12,168 / 263,516 & 66 \\
ST Gold (PMB)   & n/a              & n/a             & 356 / 1,718      & 66 \\
UD v1.2 / v1.3  & 12,543 / 204,586 & 2,002 / 25,148  & 2,077 / 25,096   & 17 \\
\bottomrule
\end{tabular}
\end{center}
\caption{\label{tab:data} Overview of the semantic tagging data (ST) and universal dependencies (UD) data.}
\end{table}

%% file: method.tex

\section{Method}

\begin{figure}[h]
    \centering
    \includegraphics[width=\textwidth]{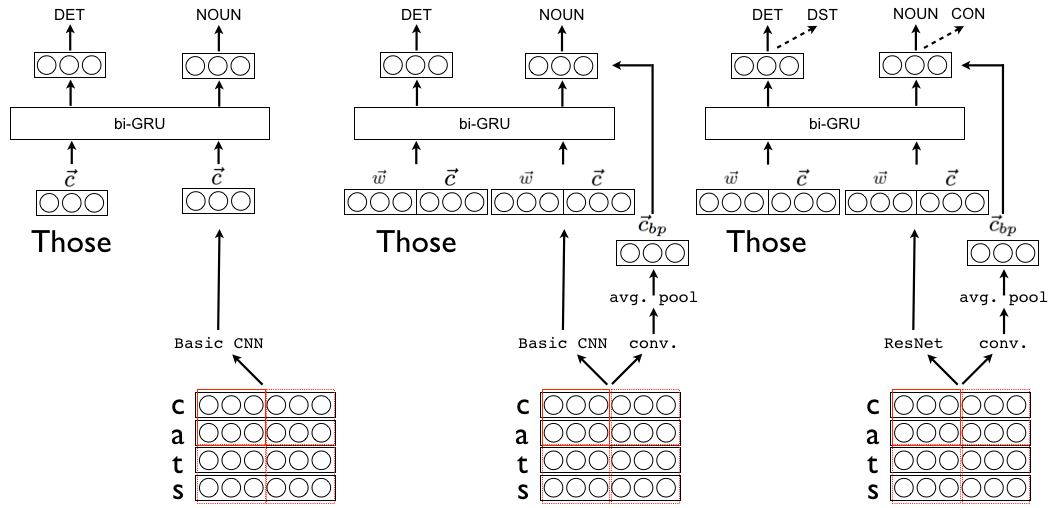}
    \caption{\label{fig:model_arch}Model architecture. Left: Architecture with basic CNN char representations ($\vec{c}$), Middle: basic CNN with char and word representations and bypass ($\vec{c}_{bp}\land\vec{w}$), Right: ResNet with auxiliary loss and residual bypass ($+${\sc aux$_{bp}$}).}
\end{figure}

Our tagger is a hierarchical deep neural network consisting of a bidirectional Gated Recurrent Unit (GRU) network at the upper level, and a Convolutional Neural Network (CNN) and/or Deep Residual Network (ResNet) at the lower level, including an optional novel residual bypass function (cf.\ Figure~\ref{fig:model_arch}).

\subsection{Gated Recurrent Unit networks}
GRUs~\cite{cho:ea:2014} are a recently introduced variant of RNNs, and are designed to prevent vanishing gradients, thus being able to cope with longer input sequences than vanilla RNNs.
GRUs are similar to the more commonly-used Long Short-Term Memory networks (LSTMs), both in purpose and implementation \cite{gru}.
A bi-directional GRU is a GRU which makes both forward and backward passes over sequences, and can therefore use both preceding and succeeding contexts to predict a tag \cite{graves:schmidhuber:2005,goldberg:primer}.
Bi-directional GRUs and LSTMs have been shown to yield high performance on several NLP tasks, such as POS tagging, named entity tagging, and chunking \cite{wang2015:unified,yang:2016,plank:2016}.
We build on previous approaches by combining bi-GRUs with character representations from a basic CNN and ResNets.

\subsection{Deep Residual Networks}
Deep Residual Networks (ResNets) are built up by stacking residual units.
A residual unit can be expressed as:
\begin{equation}
    \begin{aligned}
    &y_l = h(x_l) + \mathcal{F}(x_l,\mathcal{W}_l),\\
    &x_{l+1} = f(y_l),
    \end{aligned}
\end{equation}
\noindent where $x_l$ and $x_{l+1}$ are the input and output of the $l$-th layer, $\mathcal{W}_l$ is the weights for the $l$-th layer, and $\mathcal{F}$ is a residual function \cite{resnets:2016}, e.g., the identity function \cite{resnets:2015}, which we also use in our experiments.
ResNets can be intuitively understood by thinking of residual functions as paths through which information can propagate easily.
This means that, in every layer, a ResNet learns more complex feature combinations, which it combines with the shallower representation from the previous layer.
This architecture allows for the construction of much deeper networks.
ResNets have recently been found to yield impressive performance in image recognition tasks, with networks as deep as 1001 layers \cite{resnets:2015,resnets:2016}, and are thus an interesting and effective alternative to simply stacking layers.
In this paper we use the \textit{assymetric} variant of ResNets as described in Equation 9 in He et al.~\shortcite{resnets:2016}:
\begin{equation}
    \begin{aligned}
    &x_{l+1} = x_l + \mathcal{F}(\hat{f}(x_l), \mathcal{W}_l).
    \end{aligned}
\end{equation}

ResNets have been very recently applied in NLP to morphological reinflection \cite{robert:sigmorphon:2016}, language identification \cite{bjerva:2016:dsl}, sentiment analysis and text categorisation \cite{conneau:2016}, as well as machine translation \cite{google:nmt}.
Our work is the first to apply ResNets to NLP sequence tagging tasks.
We further contribute to the literature on ResNets by introducing a residual bypass function. The intuition is to combine both deep and shallow processing, which opens a path of easy signal propagation between lower and higher layers in the network.

\subsection{Modelling character information and residual bypass}
\label{sec:char}
Using sub-token representations instead of, or in combination with, word-level representations has recently obtained a lot of attention due to their effectiveness 
\cite{sutskever:2011,chrupala:2013,zhang:2015,chung:charactermt,gillick:2015}. 
The use of sub-token representations can be approached in several ways.
Plank et al.~\shortcite{plank:2016} and Yang et al.~\shortcite{yang:2016} use a hierarchical bi-directional RNN, first passing over characters in order to create word-level representations.
Gillick et al.~\shortcite{gillick:2015} similarly apply an LSTM-based model using byte-level information directly.
Dos Santos and Zadrozny~\shortcite{dossantos:14} construct character-based word-level representations by running a convolutional network over the character representations of each word.
All of these approaches have in common that the character-based representation is passed through the entire remainder of the network.
Our work is the first to combine the use of character-level representations with both deep processing (i.e., passing this representation through the network) and shallow processing (i.e., bypassing the network in our residual bypass function).
We achieve this by applying our novel residual bypass function to our character representations, inspired by the success of ResNets. In particular, we first apply the bypass to a CNN-based model achieving large gains over a plain CNN, and later evaluate its effectiveness in a ResNet.

A core intuition behind CNNs is the processing of an input signal in a hierarchical manner \cite{convnets:lecun,deeplearningbook}.
Taking, e.g., a 3-dimensional image ($width\times height\times depth$), the approach is typically to reduce spatial dimensions of the image while increasing depth.
This hierarchical processing allows a CNN to learn high-level features of an input, essential to image recognition tasks.
A drawback of this method, however, is that lower-level features are potentially lost in the abstraction to higher-level features.
This issue is partially countered by ResNets, as information is allowed to flow more easily between residual blocks.
However, this approach does not allow for simple and direct use of information in the network input in final layers.
To alleviate this issue, we present a residual bypass function, which can be seen as a global residual function (depicted in Figure~\ref{fig:model_arch}).
This function allows both lower-level and higher-level features to be taken directly into account in the final layers of the network.
The intuition behind using such a global residual function in NLP is that character information primarily ought to be of importance for the prediction of the current word.
Hence, allowing these representations to bypass our bi-GRU might be beneficial.
This residual bypass function is not dependent on the usage of ResNets, and can be combined with other NN architectures as in our experiments.
We define the penultimate layer of a network with $n$ layers, using a residual bypass, as follows:
\begin{equation}
    \begin{aligned}
    y_{n-1} = h(x_{n-1}) + \mathcal{F}(x_i,\mathcal{W}_i),
    \end{aligned}
\end{equation}
where $x_i$ and $\mathcal{W}_i$ are the input and weights of the $i_{th}$ layer, $\mathcal{F}$ is a residual function (in our case the identity function), and $h(x_{n-1})$ is the output of the penultimate layer.
In our experiments, we apply a residual bypass function to our convolutional character representations.

%% file: model_description.tex
\subsection{System description}

The core of our architecture consists of a bi-GRU taking an input based on words and/or characters, with an optional residual bypass as defined in subsection~\ref{sec:char}.
We experiment with a basic CNN, ResNets and our novel residual bypass function.
We also implemented a variant of the \textit{Inception} model \cite{inception:2015}, but found this to be outperformed by ResNets.
Our system is implemented in Keras using the Tensorflow backend \cite{keras,tensorflow}, and the code is available at \mbox{\url{https://github.com/bjerva/semantic-tagging}}.

We represent each sentence using both a character-based representation ($S_c$) and a word-based representation ($S_w$).
The character-based representation is a 3-dimensional matrix $S_c \in \mathbb{R}^{s \times w \times d_c}$, where $s$ is the zero-padded sentence length, $w$ is the zero-padded word length, and $d_c$ is the dimensionality of the character embeddings.
The word-based representation is a 2-dimensional matrix $S_w \in \mathbb{R}^{s \times d_w}$, where $s$ is the zero-padded sentence length and $d_w$ is the dimensionality of the word embeddings.
We use the English Polyglot embeddings \cite{polyglot} in order to initialise the word embedding layer, but also experiment with randomly initialised word embeddings.

Word embeddings are passed directly into a two-layer bi-GRU \cite{gru}.
We also experimented using a bi-LSTM.
However, we found GRUs to yield comparatively better validation data performance on semtags.
We also observe better validation data performance when running two consecutive forward and backward passes before concatenating the GRU layers, rather than concatenating after each forward/backward pass as is commonplace in NLP literature.

We use CNNs for character-level modelling.
Our basic CNN is inspired by dos Santos and Zadrozny~\shortcite{dossantos:14}, who use character-representations to produce local features around each character of a word, and combine these with a maximum pooling operation in order to create fixed-size character-level word embeddings.
The convolutions used in this manner cover a few neighbouring letters at a time, as well as the entire character vector dimension ($d_c$).
In contrast to dos Santos and Zadrozny~\shortcite{dossantos:14}, we treat a word analogously to an image.
That is to say, we see a word of $n$ characters embedded in a space with dimensionality $d_c$ as an image of dimensionality $n\times d_c$.
This view gives us additional freedom in terms of sizes of convolutional patches used, which offers more computational flexibility than using only, e.g., $4\times d_c$ convolutions.
This view is applied to all CNN variations explored in this work.


A neural network is trained with respect to some loss function, such as the cross-entropy between the predicted tag probability distribution and the gold probability distribution.
Recent work has shown that the addition of an auxiliary loss function can be beneficial to several tasks.
Cheng et al.~\shortcite{cheng:2015} use a language modelling task as an auxiliary loss, as they attempt to predict the next token while performing named entity recognition.
Plank et al.~\shortcite{plank:2016} use the log frequency of the current token as an auxiliary loss function, and find this to improve POS tagging accuracy.
Since our semantic tagging task is based on predicting fine semtags, which can be mapped to coarse semtags, we add the prediction of these coarse semtags as an auxiliary loss for the sem-tagging experiments.
Similarly, we also experiment with POS tagging, where we use the fine semtags as an auxiliary information.

\subsubsection{Hyperparameters}

All hyperparameters are tuned with respect to loss on the semtag validation set.
We use rectified linear units (ReLUs) for all activation functions \cite{relu}, and apply dropout with $p=0.1$ to both input weights and recurrent weights in the bi-GRU \cite{dropout}.
In the CNNs, we apply batch normalisation \cite{batchnorm} followed by dropout with $p=0.5$ after each layer.
In our basic CNN, we apply a $4\times8$ convolution, followed by $2\times2$ maximum pooling, followed by $4\times4$ convolution and another $2\times2$ maximum pooling.
Our ResNet has the same setup, with the addition of a residual connection.
We also experimented with using average pooling instead of maximum pooling, but this yielded lower validation data performance on the semantic tagging task.
We set both $d_c$ and $d_w$ to $64$.
All GRU layers have $100$ hidden units.
All experiments were run with early stopping monitoring validation set loss, using a maximum of 50 epochs.
We use a batch size of 500.
Optimisation is done using the ADAM algorithm \cite{adam}, with the categorical cross-entropy loss function as training objective.
The main and auxiliary loss functions have a weighting parameter, $\lambda$.
In our experiments, we weight the auxiliary loss with $\lambda=0.1$, as set on the semtag auxiliary task. 

Multi-word expressions (MWEs) are especially prominent in the semtag data, where they are annotated as single tokens.
Pre-trained word embeddings are unlikely to include entries such as `International Organization for Migration', so we apply a simple heuristic in order to avoid treating most MWEs as unknown words. In particular,
the representation of a MWE is set to the sum of the individual embeddings of each constituent word.

%% file: experiments.tex
\section{Evaluation}

We evaluate our tagger on two tasks: semantic tagging and POS tagging.
Note that the tagger is developed solely on the semantic tagging task, using the GMB silver training and validation data.
Hence, no further fine-tuning of hyperparameters for the POS tagging task is performed.
We calculate significance using bootstrap resampling \cite{efron:bootstrap}.
We manipulate the following independent variables in our experiments:
\begin{enumerate}
    \item character and word representations ($\vec{w}, \vec{c}$);
    \item residual bypass for character representations ($\vec{c}_{bp}$);
    \item convolutional representations (Basic CNN and ResNets);
    \item auxiliary loss (using coarse semtags on ST and fine semtags on UD).
\end{enumerate}

\noindent We compare our results to four baselines:
\begin{enumerate}
    \item the most frequent baseline per word (MFC), where we assign the most frequent tag for a word in the training data to that word in the test data, and unseen words get the global majority tag;
    \item the trigram statistic based TNT tagger offers a slightly tougher baseline \cite{tnt};
    \item the {\sc Bi-lstm} baseline, running the off-the-shelf state-of-the-art POS tagger for the UD dataset \cite{plank:2016} (using default parameters with pre-trained Polyglot embeddings);
    \item we use a baseline consisting of running our own system with only a {\sc Bi-gru} using word representations ($\vec{w}$), with pre-trained Polyglot embeddings.
\end{enumerate}

\subsection{Experiments on semantic tagging}
We evaluate our system on two semantic tagging (ST) datasets: our silver semtag dataset and our gold semtag dataset.
For the $+${\sc aux} condition we use coarse semtags as an auxiliary loss.
Results from these experiments are shown in Table~\ref{tab:stag_results}.

\setlength{\tabcolsep}{2.2pt}
\begin{table}[htbp]
    \small
\begin{center}
\begin{tabular}{lcccc|cccc|cccccc}
\toprule
     & \multicolumn{4}{c}{\sc Baselines}                                                & \multicolumn{4}{c}{\sc Basic CNN}       & \multicolumn{6}{c}{\sc ResNet}          \\
                     & \sc MFC & \sc TNT & \sc Bi-lstm & \sc Bi-gru & $\vec{c}$ & $\vec{c}_{bp}$ & $\vec{c}_{bp}\land\vec{w}$ & $+${\sc aux$_{bp}$}
                     & $\vec{c}$ & $\vec{c}\land\vec{w}$ & $+${\sc aux} & $\vec{c}_{bp}$ & $\vec{c}_{bp}\land\vec{w}$  & $+${\sc aux$_{bp}$} \\
\midrule
ST Silver  & 84.64 & 92.09  & 94.98 & 94.26 & 91.39  & 90.18 & 94.63 & 94.53 & 94.39 & 95.14     & 94.23 & 94.23 & \bf 95.15 & 94.58 \\
ST Gold    & 77.39 & 80.73  & 82.96 & 80.26 & 69.21  & 65.77 & 76.83 & 80.73 & 76.89 & \bf 83.64 & 74.84 & 75.84 &  82.18    & 73.73 \\
\bottomrule
\end{tabular}
\end{center}
\caption{\label{tab:stag_results}
Experiment results on semtag (ST) test sets (\% accuracy).
{\sc MFC} indicates the per-word most frequent class baseline,
{\sc TNT} indicates the TNT tagger, and
{\sc Bi-lstm} indicates the system by Plank et al.~\shortcite{plank:2016}.
{\sc Bi-gru} indicates the $\vec{w}$ only baseline.
$\vec{w}$ indicates usage of word representations,
$\vec{c}$ indicates usage of character representations, and
$\vec{c}_{bp}$ indicates usage of residual bypass of character representations.
The $+${\sc aux} column indicates the usage of an auxiliary loss.}
\end{table}

\subsection{Experiments on Part-of-Speech tagging}
We evaluate our system on v1.2 and v1.3 of the English part of the Universal Dependencies (UD) data.
We report results for POS tagging alone, comparing to commonly used baselines and prior work using LSTMs, as well as using the fine-grained semantic tags as auxiliary information.
For the $+${\sc aux} condition, we train a single joint model using a multi-task objective, with POS and ST as our two tasks.
This model is trained on the concatenation of the ST silver data with the UD data, updating the loss of the respective task of an instance in each iteration.
Hence, the weights leading to the UD softmax layer are not updated on the ST silver portion of the data, and vice-versa for the ST softmax layer on the UD portion of the data.
Results from these experiments are shown in Table~\ref{tab:pos_results}.

\begin{table}[htbp]
    \small
\begin{center}
\begin{tabular}{lcccc|cccc|cccccc}
\toprule
     & \multicolumn{4}{c}{\sc Baselines}                                                & \multicolumn{4}{c}{\sc Basic CNN}       & \multicolumn{6}{c}{\sc ResNet}          \\
                     & \sc MFC & \sc TNT & \sc Bi-lstm & \sc Bi-gru & $\vec{c}$ & $\vec{c}_{bp}$ & $\vec{c}_{bp}\land\vec{w}$ & $+${\sc aux$_{bp}$}
                     & $\vec{c}$ & $\vec{c}\land\vec{w}$ & $+${\sc aux} & $\vec{c}_{bp}$ & $\vec{c}_{bp}\land\vec{w}$  & $+${\sc aux$_{bp}$} \\
\midrule
UD v1.2         & 85.06 & 92.66  & 95.17 & 94.39 & 77.63 & 83.53 & 94.68 & 95.19 & 92.65 & 94.92 & \bf 95.71 & 92.45 & 94.73  & 95.51  \\
UD v1.3         & 85.07 & 92.69  & 95.04 & 94.32 & 77.51 & 82.89 & 94.89 & 95.34 & 92.63 & 94.88 & \bf 95.67 & 92.86 & 94.69  & 95.57  \\
\bottomrule
\end{tabular}
\end{center}
\caption{\label{tab:pos_results}
Experiment results on Universal Dependencies (UD) test sets (\% accuracy).
Adding semtags as auxiliary tags results in the best results obtained so far on English UD datasets.
}
\end{table}

%
%
%

%% file: discussion.tex
\section{Discussion}
\subsection{Performance on semantic tagging}





The overall best system is the ResNet combining both word and character representations $\vec{c} \land\vec{w}$.
It outperforms all baselines, including the recently proposed RNN-based bi-LSTM.
On the ST silver data, a significant difference ($p<0.01$) is found when comparing our best system to the strongest baseline ({\sc bi-lstm}).
On the ST gold data, we observe significant differences at the alpha values recommended by S{\o}gaard et al.~\shortcite{sogaard:2014}, with $p<0.0025$.
The residual bypass effectively helps improve the performance of the basic CNN.
However, the tagging accuracy of the CNN falls below baselines.
In addition, the large gap between gold and silver data for the CNN shows that the CNN model is more prone to overfitting, thus favouring the use of the ResNet.
Adding the coarse-grained semtags as auxiliary task only helps for the weaker CNN model.
The ResNet does not benefit from this additional information, which is already captured in the fine-grained labels.

It is especially noteworthy that the ResNet character-only system performs remarkably well, as it outperforms the {\sc Bi-gru} and TNT baselines, and is considerably better than the basic CNN.
Since performance increases further when adding in $\vec{w}$, it is clear that the character and word representations are complimentary in nature.
The high results for characters only are particularly promising for multilingual language processing, as this allows for much more compact models (see, e.g., Gillick et al.~\shortcite{gillick:2015}), which is a direction we want to explore next.

\subsection{Performance on Part-of-Speech tagging}

Our system was tuned solely on semtag data.
This is reflected in, e.g., the fact that even though our $\vec{c}\land\vec{w}$ ResNet system outperforms the Plank et al.~\shortcite{plank:2016} system on semtags, we are substantially outperformed on UD 1.2 and 1.3 in this setup.
However, adding an auxiliary loss based on our semtags markedly increases performance on POS tagging.
In this setting, our tagger outperforms the {\sc Bi-lstm} system, and results in new state-of-the-art results on both UD 1.2 ($95.71\%$ accuracy) and 1.3 ($95.67\%$ accuracy).
The difference between the {\sc Bi-lstm} system and our best system is significant at $p<0.0025$.

The fact that the semantic tags improve POS tagging performance reflects two properties of semantic tags.
Firstly, it indicates that the semantic tags carry important information which aids the prediction of POS tags.
This should come as no surprise, considering the fact that the semtags abstract over and carry more information than POS tags.
Secondly, it indicates that the new semantic tagset and released dataset are useful for downstream NLP tasks.
In this paper we show this by using semtags as an auxiliary loss.
In future work we aim to investigate the effect of introducing the semtags directly as features into the embedded input representation.

\subsection{ResNets for sequence tagging}

This work is the first to apply ResNets to NLP tagging tasks.
Our experiments show that ResNets significantly outperform standard convolutional networks on both POS tagging and sem-tagging.
ResNets allow better signal propagation and carry lower risk of overfitting, allowing for the model to capture more elaborate feature representations than in a standard CNN.


\subsection{Pre-trained embeddings}

In our main experiments, we initialised the word embedding layer with pre-trained polyglot embeddings.
We also compared this with initialising this layer from a uniform distribution over the interval $[-0.05,0.05)$.
For semantic tagging, the difference with random initialisation is negligible, with pre-trained embeddings yielding an increase in about 0.04\% accuracy.
For POS tagging, however, using pre-trained embeddings increased accuracy by almost 3 percentage points for the ResNet.

%% file: conclusions.tex
\section{Conclusions}

We introduce a semantic tagset tailored for multilingual semantic parsing.
We evaluate tagging performance using standard CNNs and the recently emerged ResNets.
ResNets are more robust and result in our best model. Combining word and
ResNet-based character representations helps to outperform state-of-the-art taggers on semantic tagging.
Coupling this with an auxiliary loss from our semantic tagset yields state-of-the-art performance on the English UD 1.2 and 1.3 POS datasets.

%% file: semtagger.bbl
\begin{thebibliography}{}

\bibitem[\protect\citename{Abadi \bgroup et al.\egroup }2016]{tensorflow}
Mart{\'{\i}}n Abadi, Ashish Agarwal, Paul Barham, Eugene Brevdo, Zhifeng Chen,
  Craig Citro, Gregory~S. Corrado, Andy Davis, Jeffrey Dean, Matthieu Devin,
  Sanjay Ghemawat, Ian~J. Goodfellow, Andrew Harp, Geoffrey Irving, Michael
  Isard, Yangqing Jia, Rafal J{\'{o}}zefowicz, Lukasz Kaiser, Manjunath Kudlur,
  Josh Levenberg, Dan Mane, Rajat Monga, Sherry Moore, Derek~Gordon Murray,
  Chris Olah, Mike Schuster, Jonathon Shlens, Benoit Steiner, Ilya Sutskever,
  Kunal Talwar, Paul~A. Tucker, Vincent Vanhoucke, Vijay Vasudevan, Fernanda~B.
  Vi{\'{e}}gas, Oriol Vinyals, Pete Warden, Martin Wattenberg, Martin Wicke,
  Yuan Yu, and Xiaoqiang Zheng.
\newblock 2016.
\newblock Tensorflow: Large-scale machine learning on heterogeneous distributed
  systems.
\newblock {\em arXiv preprint arXiv:1603.04467}.

\bibitem[\protect\citename{Al-Rfou \bgroup et al.\egroup }2013]{polyglot}
Rami Al-Rfou, Bryan Perozzi, and Steven Skiena.
\newblock 2013.
\newblock Polyglot: Distributed word representations for multilingual nlp.
\newblock {\em CoNLL-2013}.

\bibitem[\protect\citename{Berant and Liang}2014]{berant:2014}
Jonathan Berant and Percy Liang.
\newblock 2014.
\newblock Semantic parsing via paraphrasing.
\newblock In {\em ACL}, pages 1415--1425.

\bibitem[\protect\citename{Bjerva}2016]{bjerva:2016:dsl}
Johannes Bjerva.
\newblock 2016.
\newblock Byte-based language identification with deep convolutional networks.
\newblock {\em arXiv preprint arXiv:1609.09004}.

\bibitem[\protect\citename{Bos \bgroup et al.\egroup }Forthcoming]{gmb:hla}
Johan Bos, Valerio Basile, Kilian Evang, Noortje Venhuizen, and Johannes
  Bjerva.
\newblock Forthcoming.
\newblock {The Groningen Meaning Bank}.
\newblock In Nancy Ide and James Pustejovsky, editors, {\em The Handbook of
  Linguistic Annotation}. Springer, Berlin.

\bibitem[\protect\citename{Bos}2008]{step:boxer}
Johan Bos.
\newblock 2008.
\newblock {Wide-Coverage Semantic Analysis with Boxer}.
\newblock In J.~Bos and R.~Delmonte, editors, {\em Semantics in Text
  Processing. STEP 2008 Conference Proceedings}, volume~1 of {\em Research in
  Computational Semantics}, pages 277--286. College Publications.

\bibitem[\protect\citename{Brants}2000]{tnt}
Thorsten Brants.
\newblock 2000.
\newblock Tnt: a statistical part-of-speech tagger.
\newblock In {\em Proceedings of the sixth conference on Applied natural
  language processing}, pages 224--231. Association for Computational
  Linguistics.

\bibitem[\protect\citename{Butler}2010]{Butler:2010}
Alastair Butler.
\newblock 2010.
\newblock {\em The Semantics of Grammatical Dependencies}, volume~23.
\newblock Emerald Group Publishing Limited.

\bibitem[\protect\citename{Cheng \bgroup et al.\egroup }2015]{cheng:2015}
Hao Cheng, Hao Fang, and Mari Ostendorf.
\newblock 2015.
\newblock Open-domain name error detection using a multi-task rnn.
\newblock In {\em EMNLP}.

\bibitem[\protect\citename{Cho \bgroup et al.\egroup }2014]{cho:ea:2014}
Kyunghyun Cho, Bart Van~Merri{\"e}nboer, Caglar Gulcehre, Dzmitry Bahdanau,
  Fethi Bougares, Holger Schwenk, and Yoshua Bengio.
\newblock 2014.
\newblock Learning phrase representations using rnn encoder-decoder for
  statistical machine translation.
\newblock In {\em EMNLP}.

\bibitem[\protect\citename{Chollet}2015]{keras}
Fran\c{c}ois Chollet.
\newblock 2015.
\newblock Keras.
\newblock \url{https://github.com/fchollet/keras}.

\bibitem[\protect\citename{Chrupa{\l}a}2013]{chrupala:2013}
Grzegorz Chrupa{\l}a.
\newblock 2013.
\newblock Text segmentation with character-level text embeddings.
\newblock In {\em Workshop on Deep Learning for Audio, Speech and Language
  Processing, ICML}.

\bibitem[\protect\citename{Chung \bgroup et al.\egroup }2014]{gru}
Junyoung Chung, Caglar Gulcehre, KyungHyun Cho, and Yoshua Bengio.
\newblock 2014.
\newblock Empirical evaluation of gated recurrent neural networks on sequence
  modeling.
\newblock {\em arXiv preprint arXiv:1412.3555}.

\bibitem[\protect\citename{Chung \bgroup et al.\egroup
  }2016]{chung:charactermt}
Junyoung Chung, Kyunghyun Cho, and Yoshua Bengio.
\newblock 2016.
\newblock A character-level decoder without explicit segmentation for neural
  machine translation.
\newblock {\em Procedings of ACL 2016, arXiv preprint arXiv:1603.06147}.

\bibitem[\protect\citename{Conneau \bgroup et al.\egroup }2016]{conneau:2016}
Alexis Conneau, Holger Schwenk, Lo{\"\i}c Barrault, and Yann Lecun.
\newblock 2016.
\newblock Very deep convolutional networks for natural language processing.
\newblock {\em arXiv preprint arXiv:1606.01781}.

\bibitem[\protect\citename{Copestake \bgroup et al.\egroup }2005]{mrs}
Ann Copestake, Dan Flickinger, Ivan Sag, and Carl Pollard.
\newblock 2005.
\newblock Minimal recursion semantics: An introduction.
\newblock {\em Journal of Research on Language and Computation},
  3(2--3):281--332.

\bibitem[\protect\citename{Curran \bgroup et al.\egroup }2007]{candcboxer:2007}
James Curran, Stephen Clark, and Johan Bos.
\newblock 2007.
\newblock {Linguistically Motivated Large-Scale NLP with C\&C and Boxer}.
\newblock In {\em Procedings of ACL 2007}, pages 33--36, Prague, Czech
  Republic.

\bibitem[\protect\citename{dos Santos and Zadrozny}2014]{dossantos:14}
C{\'\i}cero~Nogueira dos Santos and Bianca Zadrozny.
\newblock 2014.
\newblock Learning character-level representations for part-of-speech tagging.
\newblock In {\em ICML}, pages 1818--1826.

\bibitem[\protect\citename{Efron and Tibshirani}1994]{efron:bootstrap}
Bradley Efron and Robert~J Tibshirani.
\newblock 1994.
\newblock {\em An introduction to the bootstrap}.
\newblock CRC press.

\bibitem[\protect\citename{Evang \bgroup et al.\egroup }2013]{elephant}
Kilian Evang, Valerio Basile, Grzegorz Chrupa{\l}a, and Johan Bos.
\newblock 2013.
\newblock {Elephant}: Sequence labeling for word and sentence segmentation.
\newblock In {\em EMNLP}, pages 1422--1426.

\bibitem[\protect\citename{Gillick \bgroup et al.\egroup }2015]{gillick:2015}
Dan Gillick, Cliff Brunk, Oriol Vinyals, and Amarnag Subramanya.
\newblock 2015.
\newblock Multilingual language processing from bytes.
\newblock {\em arXiv preprint arXiv:1512.00103}.

\bibitem[\protect\citename{Goldberg}2015]{goldberg:primer}
Yoav Goldberg.
\newblock 2015.
\newblock A primer on neural network models for natural language processing.
\newblock {\em arXiv preprint arXiv:1510.00726}.

\bibitem[\protect\citename{Goodfellow \bgroup et al.\egroup
  }2016]{deeplearningbook}
Ian Goodfellow, Yoshua Bengio, and Aaron Courville.
\newblock 2016.
\newblock Deep learning.
\newblock \mbox{\url{http://www.deeplearningbook.org}}.
\newblock Book in preparation for MIT Press.

\bibitem[\protect\citename{Graves and
  Schmidhuber}2005]{graves:schmidhuber:2005}
Alex Graves and J{\"u}rgen Schmidhuber.
\newblock 2005.
\newblock Framewise phoneme classification with bidirectional lstm and other
  neural network architectures.
\newblock {\em Neural Networks}, 18(5):602--610.

\bibitem[\protect\citename{He \bgroup et al.\egroup }2015]{resnets:2015}
Kaiming He, Xiangyu Zhang, Shaoqing Ren, and Jian Sun.
\newblock 2015.
\newblock Deep residual learning for image recognition.
\newblock {\em arXiv preprint arXiv:1512.03385}.

\bibitem[\protect\citename{He \bgroup et al.\egroup }2016]{resnets:2016}
Kaiming He, Xiangyu Zhang, Shaoqing Ren, and Jian Sun.
\newblock 2016.
\newblock Identity mappings in deep residual networks.
\newblock {\em arXiv preprint arXiv:1603.05027}.

\bibitem[\protect\citename{Ioffe and Szegedy}2015]{batchnorm}
Sergey Ioffe and Christian Szegedy.
\newblock 2015.
\newblock Batch normalization: Accelerating deep network training by reducing
  internal covariate shift.
\newblock {\em arXiv preprint arXiv:1502.03167}.

\bibitem[\protect\citename{Kingma and Ba}2014]{adam}
Diederik Kingma and Jimmy Ba.
\newblock 2014.
\newblock Adam: A method for stochastic optimization.
\newblock {\em arXiv preprint arXiv:1412.6980}.

\bibitem[\protect\citename{LeCun \bgroup et al.\egroup }1998]{convnets:lecun}
Yann LeCun, L{\'e}on Bottou, Yoshua Bengio, and Patrick Haffner.
\newblock 1998.
\newblock Gradient-based learning applied to document recognition.
\newblock {\em Proceedings of the IEEE}, 86(11):2278--2324.

\bibitem[\protect\citename{Marcus \bgroup et al.\egroup }1993]{ptb}
M.P. Marcus, B.~Santorini, and M.A. Marcinkiewicz.
\newblock 1993.
\newblock {Building a Large Annotated Corpus of English: The Penn Treebank}.
\newblock {\em Computational Linguistics}, 19(2):313--330.

\bibitem[\protect\citename{Nair and Hinton}2010]{relu}
Vinod Nair and Geoffrey~E Hinton.
\newblock 2010.
\newblock Rectified linear units improve restricted boltzmann machines.
\newblock In {\em Proceedings of the 27th International Conference on Machine
  Learning (ICML-10)}, pages 807--814.

\bibitem[\protect\citename{Nivre \bgroup et al.\egroup }2016]{nivre:2016}
Joakim Nivre, Marie-Catherine de~Marneffe, Filip Ginter, Yoav Goldberg, Jan
  Hajic, Christopher~D Manning, Ryan McDonald, Slav Petrov, Sampo Pyysalo,
  Natalia Silveira, et~al.
\newblock 2016.
\newblock Universal dependencies v1: A multilingual treebank collection.
\newblock In {\em Proceedings of the 10th International Conference on Language
  Resources and Evaluation (LREC 2016)}.

\bibitem[\protect\citename{\"Ostling}2016]{robert:sigmorphon:2016}
Robert \"Ostling.
\newblock 2016.
\newblock Morphological reinflection with convolutional neural networks.
\newblock In {\em {Proceedings of the 2016 Meeting of SIGMORPHON}}, Berlin,
  Germany. Association for Computational Linguistics.

\bibitem[\protect\citename{Plank \bgroup et al.\egroup }2016]{plank:2016}
Barbara Plank, Anders S{\o}gaard, and Yoav Goldberg.
\newblock 2016.
\newblock Multilingual part-of-speech tagging with bidirectional long
  short-term memory models and auxiliary loss.
\newblock In {\em Proceedings of ACL 2016, arXiv preprint arXiv:1604.05529}.

\bibitem[\protect\citename{Pradhan \bgroup et al.\egroup }2004]{pradhan:2004}
Sameer~S Pradhan, Wayne Ward, Kadri Hacioglu, James~H Martin, and Daniel
  Jurafsky.
\newblock 2004.
\newblock Shallow semantic parsing using support vector machines.
\newblock In {\em HLT-NAACL}, pages 233--240.

\bibitem[\protect\citename{S{\o}gaard \bgroup et al.\egroup
  }2014]{sogaard:2014}
Anders S{\o}gaard, Anders Johannsen, Barbara Plank, Dirk Hovy, and Hector
  Martinez.
\newblock 2014.
\newblock What’s in a p-value in nlp?
\newblock In {\em CoNLL-2014}.

\bibitem[\protect\citename{Srivastava \bgroup et al.\egroup }2014]{dropout}
Nitish Srivastava, Geoffrey~E Hinton, Alex Krizhevsky, Ilya Sutskever, and
  Ruslan Salakhutdinov.
\newblock 2014.
\newblock Dropout: a simple way to prevent neural networks from overfitting.
\newblock {\em Journal of Machine Learning Research}, 15(1):1929--1958.

\bibitem[\protect\citename{Sutskever \bgroup et al.\egroup
  }2011]{sutskever:2011}
Ilya Sutskever, James Martens, and Geoffrey~E Hinton.
\newblock 2011.
\newblock Generating text with recurrent neural networks.
\newblock In {\em Proceedings of the 28th International Conference on Machine
  Learning (ICML-11)}, pages 1017--1024.

\bibitem[\protect\citename{Szegedy \bgroup et al.\egroup }2015]{inception:2015}
Christian Szegedy, Wei Liu, Yangqing Jia, Pierre Sermanet, Scott Reed, Dragomir
  Anguelov, Dumitru Erhan, Vincent Vanhoucke, and Andrew Rabinovich.
\newblock 2015.
\newblock Going deeper with convolutions.
\newblock In {\em Proceedings of the IEEE Conference on Computer Vision and
  Pattern Recognition}, pages 1--9.

\bibitem[\protect\citename{Wang \bgroup et al.\egroup }2015]{wang2015:unified}
Peilu Wang, Yao Qian, Frank~K Soong, Lei He, and Hai Zhao.
\newblock 2015.
\newblock A unified tagging solution: Bidirectional lstm recurrent neural
  network with word embedding.
\newblock {\em arXiv preprint arXiv:1511.00215}.

\bibitem[\protect\citename{Wu \bgroup et al.\egroup }2016]{google:nmt}
Yonghui Wu, Mike Schuster, Zhifeng Chen, Quoc~V Le, Mohammad Norouzi, Wolfgang
  Macherey, Maxim Krikun, Yuan Cao, Qin Gao, Klaus Macherey, et~al.
\newblock 2016.
\newblock Google's neural machine translation system: Bridging the gap between
  human and machine translation.
\newblock {\em arXiv preprint arXiv:1609.08144}.

\bibitem[\protect\citename{Yang \bgroup et al.\egroup }2016]{yang:2016}
Zhilin Yang, Ruslan Salakhutdinov, and William Cohen.
\newblock 2016.
\newblock Multi-task cross-lingual sequence tagging from scratch.
\newblock {\em arXiv preprint arXiv:1603.06270}.

\bibitem[\protect\citename{Zhang \bgroup et al.\egroup }2015]{zhang:2015}
Xiang Zhang, Junbo Zhao, and Yann LeCun.
\newblock 2015.
\newblock Character-level convolutional networks for text classification.
\newblock In {\em Advances in Neural Information Processing Systems}, pages
  649--657.

\end{thebibliography}
